\documentclass[10pt,twocolumn,letterpaper]{article}

\usepackage[table]{xcolor}
\usepackage[utf8]{inputenc}
\usepackage{cvpr}
\usepackage{times}
\usepackage{epsfig}
\usepackage{graphicx}
\usepackage{amsmath}
\usepackage{amssymb}
\usepackage{pgfplots}
\usepackage{multirow}
\usepackage{booktabs}
\usepackage{hhline}
\usepackage{subcaption}
\usepackage{eucal}

\tikzset{cross/.style={cross out, draw=black, minimum size=2*(#1-\pgflinewidth), inner sep=0pt, outer sep=0pt},
cross/.default={5pt}}

\definecolor{rowblue}{RGB}{220,230,240}
\pgfplotsset{compat=1.14}

\usepackage[pagebackref=true,breaklinks=true,letterpaper=true,colorlinks,bookmarks=false]{hyperref}

\cvprfinalcopy 

\def\cvprPaperID{6144} 

\ifcvprfinal\fi
\begin{document}

\title{Context-aware Human Motion Prediction}

\author{Enric Corona \and
Albert Pumarola \and
Guillem Aleny\`a \and
Francesc Moreno-Noguer \and
\\ Institut de Robòtica i Informàtica Industrial, CSIC-UPC, 08028, Barcelona, Spain
\\ \{ecorona, apumarola, galenya, fmoreno\}@iri.upc.edu
}

\maketitle
\thispagestyle{empty}

\begin{abstract}
The problem of predicting human motion given a sequence of past observations is at the core of many applications in robotics and computer vision. Current state-of-the-art  formulate this problem as a sequence-to-sequence task, in which a historical of 3D skeletons feeds a Recurrent Neural Network (RNN) that predicts future movements, typically in the order of 1 to 2 seconds.  However, one aspect that has been obviated so far, is the fact that human motion is inherently driven by interactions with objects and/or other humans in the environment.

In this paper, we explore this scenario using  a novel context-aware  motion prediction architecture. We use a semantic-graph model where the nodes parameterize the human and objects in the scene and the edges their mutual interactions. These interactions are iteratively learned through a graph attention layer, fed with the past observations, which now include both object and human body motions.  Once this semantic graph is learned, we inject it to a standard RNN to predict future movements  of the human/s and object/s. We consider two variants of our architecture, either freezing the contextual interactions in the future of updating them. 
A thorough evaluation in the  ``Whole-Body Human Motion Database''~\cite{koppula2013learning} shows that in both cases, our context-aware networks clearly outperform baselines in which the context information is not considered. 

\end{abstract}

\vspace{-0.1cm}
\section{Introduction}

\begin{figure*}[t!]
\includegraphics[width=\textwidth]{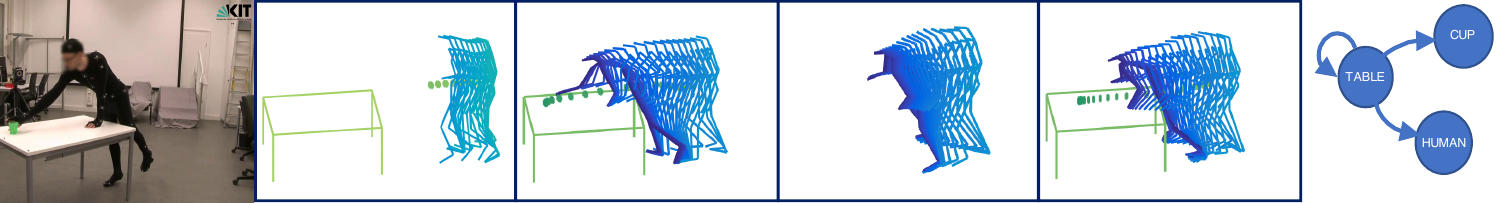}\\
\put(38, 18){\footnotesize{(a)}}
\put(125, 18){\footnotesize{(b)}}
\put(210, 18){\footnotesize{(c)}}
\put(296, 18){\footnotesize{(d)}}
\put(385, 18){\footnotesize{(e)}}
\put(460, 18){\footnotesize{(f)}}
\vspace{-8mm}
\caption{\small{{\bf Context-aware human motion prediction.} (a) Sample image  of a sequence with a person placing a cup on a table. This image is  shown solely for illustrative purposes,  our approach only relies on positional data. (b) Past observations of all elements of the scene, the person, the cup and the table. (c) Ground truth future movements. (d) Human motion predicted using~\cite{martinez2017human}, consisting of an RNN that is agnostic of the context information. Note that there is a large gap with the ground truth, especially for the final frames of the sequence. (e) Cup and human motion prediction obtained with our context-aware model. While the arm of the person is not fully extended, the forecasted motion highly resembles the ground truth. Interestingly, the interaction with the table also helps to set the motion boundaries. (f) Main interactions that are learned with our approach in which dominates the influence of the table over both the cup and the person.}}
\label{fig:intro}
\vspace{-0.2cm}

\end{figure*}
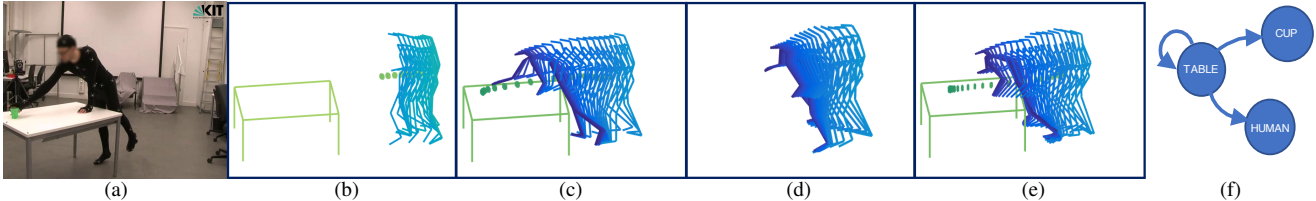

The ability to predict and anticipate  future human motion based on past observations is essential for interacting with other people and the world around us. While this seems a trivial task for a person, it  involves multiple sensory modalities and complex semantic understanding  of the environment and the relations between all objects in it. Modeling and transferring this kind of knowledge to   autonomous agents would have a major impact in many different fields,  mainly  in  human-robot interaction~\cite{koppula2016anticipating} and  autonomous driving~\cite{paden2016survey}, but also in motion generation for computer graphics animation~\cite{kovar2008motion} or image understanding~\cite{chen2018graph}.

The explosion of deep learning, combined with  large-scale datasets of human motion such as Human3.6M~\cite{h3.6m} or the CMU motion capture dataset~\cite{cmu_motion_database}, has  led to a significant amount of recent literature that tackles the problem of forecasting 3D human motion from past observations~\cite{fragkiadaki2015recurrent, jain2016structural, martinez2017human,gui2018adversarial,barsoum2018hp, li2018transferable, ghosh2017learning, Mao_2019_ICCV,kim2019instance,zhang2019predicting}.  These algorithms typically formulate the problem as sequence-to-sequence task, in which past observations  represented    3D skeleton data are injected to a Recurrent Neural Network (RNN) which then predicts movements in the near future  (less than 2 seconds). 

Nevertheless, while promising results have been achieved, we argue that the standard definition of the problem used so far lacks an important factor, which is the influence of the rest of the environment on the  movement of the person. For instance, if a person is  carrying  a box, the configuration  of the body arms and legs will be highly constrained by the 3D position of that box. Discovering such interrelations between the person and the object/s of the context (or another person he/she is interacting with), and how these interrelations constrain the body motion, is the principal motivation of this paper.

In order to explore this new paradigm, we devise a  context-aware motion prediction architecture, that models the interactions between all objects of the scene and the human using a directed semantic graph. The   nodes of this graph represent the state of the person and objects (\eg positional features) and the edges their mutual interactions. These interactions are iteratively learned with the past observations of the human and objects motion and  fed into a standard RNN which is then responsible for predicting the future movement of all elements in the scene (for both rigid objects and non-rigid human skeletons). Additionally, we     propose a variant of this model that also predicts the evolution of the adjacency matrix representing the interaction between the elements of the scene.

Presumably, one of the reasons why current state-of-the-art has not considered an scenario like ours is because all methods are trained and evaluated on benchmarks (mostly the aforementioned Human3.6M dataset~\cite{h3.6m}) annotated only with  human motion. In this paper, we thoroughly evaluate our approach in the ``Whole-Body Human Motion Database''~\cite{koppula2013learning}, that contains about 200 videos of people performing several tasks and interacting with objects. This dataset is annotated with MoCap data for the humans and rigid displacement for the rest of objects, being thus, a perfect benchmark to validate our ideas. We also evaluate our method in the CMU MoCap database~\cite{cmu_motion_database} with only two people being tracked.
The results obtained in both datasets show that our methodology is able to  accurately predict   the future motion of people and objects while simultaneously learning very coherent interaction relations. Additionally, all context-aware versions, clearly outperform the baselines which uniquely rely on human past observations of the human (see Fig.~\ref{fig:intro}).
Since all previous works evaluate their methods using past observations of ground truth skeletons, we finally discuss the applicability of state-of-art motion prediction methods, with an ablation study of our models and baselines when considering noisy observations.

\begin{figure*}
\vspace{-0.5em}
\centering
  \includegraphics[width=\linewidth]{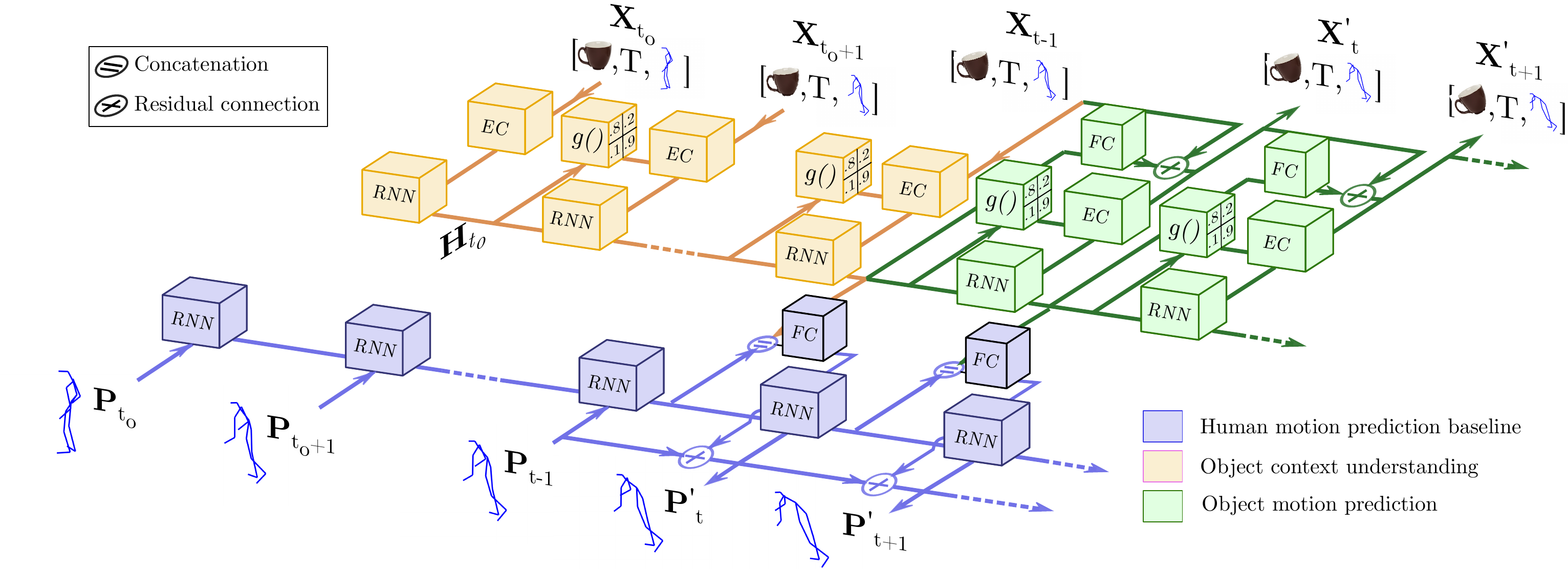}
  \vspace{-0.7cm}
  \caption{\small{{\bf Overview of our context-aware motion prediction model.} The blue branch represents a basic RNN that encodes past poses and decodes future human motion using a residual layer~\cite{martinez2017human}. The upper branch corresponds to an RNN that encodes the contextual representation for each  object in the scene. This branch  contains two modules (depicted in brown and green).  In brown, the past object position, class, and human joints are used to predict interactions and context feature vectors. The node corresponding to the human   context representation is then used in conjunction with the human motion hidden state, to predict human motion. In green, the model is extended to predict motion of all observed objects. Best viewed in color.}}
  \label{fig:model_diagram}
  \vspace{-0.2cm}
\end{figure*}

\vspace{-1mm}
\section{Related work}
\noindent{\bf Human motion prediction.}
Since the release of  large-scale MoCap datasets~\cite{pumarola20193dpeople, h3.6m, koppula2013learning}, there has been a growing interest  in the problem of estimating 3D human pose  from single images~\cite{bogo2016keep, pumarola20193dpeople, pavlakos2017coarse, wang2014robust, simo2012single, simo2013joint, Moreno-Noguer_2017_CVPR, simo2017humanpose}. More recently, the community is focusing in predicting 3D human motion from past observations. 
Most approaches build upon  RNNs~\cite{fragkiadaki2015recurrent, martinez2017human, gui2018adversarial, barsoum2018hp, pavllo2018quaternet, Aksan_2019_ICCV} that encode historical motion of the human and predict the future configuration that minimizes different sort of losses. Martinez~\etal~\cite{martinez2017human}, for instance,  minimize the L2 distance and provide one of the baselines in our work. This work also compares against a zero-velocity baseline, which despite steadily predicting the last observed frame, yields very reasonable results under the L2 metric. This phenomenon has been recently discussed by Ruiz~\etal~\cite{ruiz2019human}, that argue that L2 distance is not an appropriate metric to capture the actual distribution of human motion, and that a network trained using only this metric is prone to converge to a mean body pose. To better capture real distributions of   human movement, recent approaches  use  adversarial networks~\cite{goodfellow2014generative, arjovsky2017wasserstein} in combination with geometric losses~\cite{barsoum2018hp, gui2018adversarial, ruiz2019human, kundu2018bihmp}.

There exist alternative approaches other than RNNs. For instance, Jain~\etal~\cite{jain2016structural} consider a hand-crafted spatial-temporal graph adapted to the skeleton shape.  Li~\etal~\cite{li2018transferable} use Convolutional Neural Networks to encode and decode skeleton sequences instead of RNNs.

All  methods described in this section formulate the human prediction problem without considering the context information. In this paper, we aim to fill this gap.


\vspace{1mm}
\noindent{\bf Rigid 3D object motion prediction.}
While there is a vast amount of works on 3D object reconstruction~\cite{pumarola2020c, groueix2018atlasnet, mandikal2019dense},  detection~\cite{chen20153d, grabner20183d, corona2018pose} and tracking~\cite{chang2019argoverse, baser2019fantrack}, only very few approaches address the problem of predicting future rigid motion~\cite{byravan2017se3, xie2019rigid, kratzer2019motion, vijayanarasimhan2017sfm}. Among these, it is worth to mention Byravan~\etal~\cite{byravan2017se3}, that predict the  future 3D pose given an image of an object and the action being applied to it. In our case, the action applied to each object is implicitly encoded in the previous observations. 



\vspace{1mm}
\noindent{\bf Human-Object Interaction (HOI).}
Even though our work does not aim to identify Human-Object relationships, we have been inspired by a few papers on this topic. The standard  formulation of the problem consists in representing   an image with several detected objects and people as a graph encoding the context~\cite{herzig2018mapping, newell2017pixels, qi2018learning, ma2018attend, gkioxari2015actions}, or some other structured representation~\cite{li2017situation, yatskar2016situation, corona2020ganhand}. 
The most recent approaches~\cite{li2018transferable,qi2018learning, herzig2018mapping} extract features of the detected entities using some image-based classification CNNs. Then, they compare pairs of features to predict their mutual interaction. Qi~\etal~\cite{qi2018learning} refine the representations and predicted interactions   in a recursive manner. In this work, we use a similar idea to progressively refine the estimation of the interactions between objects.

\vspace{1mm}
\noindent{\bf Graph-based context reasoning.} A few works   leverage  context information  to boost the performance of different tasks\cite{parisot2018disease, liu2018structure, Hu_2018_CVPR, lea2016segmental, ni2016progressively}.   Graph Convolutional Networks (GCNs)~\cite{GCNs} were recently proposed for improved semi-supervised classification. Jain~\etal~\cite{jain2016structural} used Structural RNNs to model spatio-temporal graphs. Wang~\etal~\cite{wang2018videos} propose to use GCNs, in which the interactions between objects depend on the intersection over union of their detected bounding boxes. Chen~\etal~\cite{chen2018graph} introduce an approach for image segmentation in which features from a 2D image coordinate space are represented in a graph reasoning space.

\vspace{-0.1cm}
\section{Problem formulation}
\vspace{-0.1cm}

Recent methods for human motion prediction consist of a model $\mathcal{M}$, typically a deep neural network, that encodes motion from time $t_o$ until $t-1$. The goal is then to   predict future human motion until $t_f$, namely $ P_{t:t_f}$, where $P$ stands for the human pose represented by  3D joint coordinates. Previous approaches have formulated the problem as $\mathcal{M}:P_{t_o:t-1}\rightarrow P_{t:t_f}$, \ie future motion is estimated only from past observations. In this paper, we conjecture that future motion is also driven by the context and the action the human is performing. We therefore consider other objects $O$ of type $T$ in the scene with which the human may interact. The objects can be other people or any object in the scene.  We will design our approach to be able to predict the motion of such objects of the context.

Additionally, the influence that objects will have in the future motion of other objects is unclear. Thus, we  also aim to build a model that learns these interactions in an unsupervised manner. Considering all this, we reformulate our problem as the estimation of the  following mapping:
\begin{equation}
\mathcal{M}: \{P_{t_o:t-1}, O_{t_o:t-1}, T\} \rightarrow \{P_{t:t_f}, O_{t:t_f}, I_{t_o:t_f}\}\;, 
\end{equation}
\vspace{-0.1cm}
where $I$ corresponds to  the predicted interactions.

\section{Approach}
\vspace{-0.1cm}

Figure~\ref{fig:model_diagram} shows the main architecture used in this work. It consists of two branches that separately process human motion and object relationships. We use the latter to obtain a representation for all the observed entities, including the human, which we then use to predict both human and object motion prediction. We  next describe these two branches.

\subsection{Human motion branch}
This branch  builds  upon the RNN network proposed by Martinez~\etal~\cite{martinez2017human}.  This model, depicted in blue in Figure~\ref{fig:model_diagram}, is based on a residual architecture~\cite{he2016deep} that, at each step, uses a fully connected layer to predict the velocity of the body joints. As in a typical sequence-to-sequence network, the predictions are fed to the next step.

\subsection{Context branch}\label{sec:context_branch}
The context information is represented using a directed graph structure  where each node denotes an object or person. We then store a state for each entity and frame, encoding context information relevant to each node. These states are iteratively refined as new observations are processed.

\vspace{1mm}
\noindent{\bf Object representation.} At each frame $t$, we define a  matrix  $X_t \ \in \mathbb{R}^{N \times F_0} = \left[O_t, T_t, P_t\right]$ that gathers the representation of all $N$ nodes. $F_0$ is the length of the state vector of each node. This state vector  contains the object 3D bounding box $O_t$, their object type $T$ as a one-hot vector, and the joints of the person $P_t$. If the node does not correspond to a person, the joints in the representation are set to a zero vector of same size. The
object type helps to identify the task the human is performing and the motion defined for that task. 
By doing this, we aim to capture the semantic difference between the motion of a person when handling a knife or when using a whisk.

\vspace{1mm}
\noindent{\bf Modelling contextual object representations.} Recent works on Graph Convolutional Networks (GCNs)~\cite{GCNs} have shown very promising results in a variety of problems requiring the manipulation of graph-structured data. In GCNs, a feature vector of a certain node $R_i$ is expressed as a function of other nodes $x$, as $R_i = \sigma (\sum^{N}_j  \tilde{A}_{ij} W x_j )$, where $W$ are trainable weights, $\sigma$ is an activation and $N$ the number of nodes of the graph connected to the $i$-th node. $\tilde{A} \in \mathbb{R}^{N \times N}$ is a  normalized weighted adjacency matrix  that defines interactions between nodes. 

Graph Attention Networks (GATs)~\cite{velivckovic2017graph} have been proposed as an extension of GCNs, and introduce an attention model on every graph node. In this paper we also investigate the use of Edge Convolutions~\cite{EdgeConvs}, which are indeed very similar to GATs. In ECs the update rule for  a feature vector of each entity considers the   representations of other relevant objects as follows: 

\vspace{-0.2cm}
\begin{equation}\label{eq:edgeconv}
	R_i = \sigma ( \sum^{N}_j  \tilde{A}_{ij} W [x_i; x_i - x_j] ).
\end{equation}
The intuition behind this equation is that $x_i$ encodes a global representation of the node, while $x_i - x_j$ provides   local   information. EC proposes combining both types of information in an asymmetric graph function. 

We keep track of the context representations during all observations through a second RNN. Each node on the scene has a hidden state $H$ that is   updated every frame $t$:
\vspace{-0.1cm}
\begin{equation}
H^{t+1}_i = \textrm{RNN}( R_i^t, H^{t}_i).
\end{equation}

\noindent{\bf Learning interactions.} As we shall see in the experimental section, we initially evaluate a simplified version of our Context-RNN (C-RNN) that uses a  heuristic to define the adjacency matrices, setting $A_{ij} = 1$  if the center of gravity of objects $i$ and $j$ is closer than 1 meter.  

In practice, interactions between entities are not known a   priory, and furthermore, they change over time. Our goal is to automatically learn these changing interactions with no supervision. For this purpose we devise an iterative process in which, for the first frame, we set $A$ to a diagonal matrix, \ie  $ \tilde{A}_{t_0} = I_N$, meaning that the initial hidden representation of every object depends only on itself. We then predict the value of the interaction between two objects given the hidden state of both. We consider asymmetric weighted adjacency matrices, that for a frame $t$ are estimated as:
\begin{equation}
A_{ij}^t = g(H_i^t, H_i^t - H_j^t),
\end{equation}
with similar structure as in Eq.~\ref{eq:edgeconv}. The function $g$ represents the output of a neural network layer, in our case a fully connected. We normalize the interactions for each node using a   Softmax function, which we shall denote   $\tilde{A}$.

Intuitively, we can consider this as a complete graph, where a graph attention mechanism~\cite{velivckovic2017graph} decides on the strength of interactions based on past observations. Note that while existing works typically use binary adjacency matrices from ground truth relationships~\cite{GCNs}, spatial assumptions~\cite{wang2018videos} or K-NN on node representations~\cite{EdgeConvs}, in this work we consider a differentiable continuous space of interactions, learned using back-propagation.
In the rest of the paper we will denote the models that learn interactions with the suffix ``-LI'' (\eg C-RNN+LI).

\vspace{1mm} \noindent{\bf Object motion prediction.} We propose two methods that exploit context at different levels. First, in the  blue+brown modules of Fig.~\ref{fig:model_diagram}, we consider a model that reasons about the past context observations and iteratively improves hidden representations.
The refined context representation of the human node is concatenated to the baseline branch (in blue) representation at every time step, and used by a fully connected layer to predict human velocity in that step. This is followed by a residual layer that yields   skeleton poses.

Our second approach consists of the complete model depicted in Fig.~\ref{fig:model_diagram} which, apart from past context, predicts object motion for all objects using a residual fully connected layer on each object hidden state. Analogous to the human motion branch, the  predicted positions are forwarded to the next step,  allowing  to extend the context analysis into the future. The joints in the feature representations for those nodes describing  people are also updated with the joint predictions of the human branch.

Additionally, when tracking  several people, the human motion branch is repeated for each of them, and the model provides complete future motion for all available entities. In the rest of the document, we will denote the models that predict object motion with the suffix ``-OPM".

\begin{table*}
\vspace{-0.5em}
\centering
\footnotesize
\rowcolors{1}{white}{rowblue}
\resizebox{\textwidth}{!}{%
\begin{tabular}{ccccc|cccc|cccc|cccc|cccc}
\hline
Human motion  & \multicolumn{4}{c}{Passing objects}                                                                                                                                                                                                                & \multicolumn{4}{c}{Grasping objects}                                                                                                                                                                                                    & \multicolumn{4}{c}{Cutting food}                                                                                                                                                                                                               & \multicolumn{4}{c}{Mixing objects}                                                                                                                                                                                                             & \multicolumn{4}{c}{Cooking}                                                                                                        
 \\ \hline
\multicolumn{1}{c|}{Time (s)}                                                                  & 0.5 & 1 & 1.5 & 2 & 0.5 &1 & 1.5 & 2 & 0.5 & 1 & 1.5 & 2 & 0.5 & 1 & 1.5 & 2 & 0.5 & 1 & 1.5 & 2 \\  \hline
\multicolumn{1}{c|}{ZV\cite{martinez2017human}}               & \textbf{34}                                                      & \textbf{81}                                                        & \underline{120}                                                       & \underline{153}                                                       & 89                                                      & 222                                                       & 333                                                      & 421                                                      & 54                                                     & 132                                                      & 198                                                      & 258                                                      & 102                                                    & 262                                                      & 396                                                      & 495 & \textbf{24} & \textbf{53} & 70 & 80 \\ 
\multicolumn{1}{c|}{RNN\cite{martinez2017human}}                & 50                                                      & 99                                                        & 132                                                       & 162                                                       & 82                                                     & 158                                                       & 211                                                      & 254                                                      & \underline{48}                                                     & 103                                                      & 140                                                      & 180                                                      & \underline{68}                                                     & \underline{135}                                                      & \underline{190}                                                      & 226                                         &27&\underline{54}&\underline{65}&71 \\ 
\multicolumn{1}{c|}{QuaterNet} & 62 & 145 & 211 & 267 & 208 & 209 & 248 & 292 & 87 & 211 & 308 & 389 & 192 & 237 & 296 & 345 & 39 & 87 & 121 & 144\\ \hline

\multicolumn{1}{c|}{\textit{C-RNN}} & 47 & 102 & 141 & 177 & 76 & \underline{149} & 203 & 247 & 49 & 100 & 124 & 158 & 70 & 158 & 214 & 247 & \underline{26} & \textbf{53} & \textbf{63} & \underline{69} \\

\multicolumn{1}{c|}{\textit{C-RNN+OMP}} & 53 & 99 & 127 & 155 & 128 & 154 & \underline{197} & \underline{239} & 49 & 96 & \underline{121 }& \underline{149} & \textbf{61} & \textbf{127} & \textbf{168} & \textbf{199} & 29 & 55 & \underline{65} & 70 \\ \hline

\multicolumn{1}{c|}{\textit{C-RNN+LI}}      & \underline{43}                                                      & \underline{89}                                                        & \underline{117}                                                       & \textbf{142}                                                       & \textbf{72}                                                      & \textbf{141}                                                       & \textbf{188}                                                      & \textbf{230}                                                      & \textbf{47}                                                     & \textbf{92}                                                       & \textbf{117}                                                      & \textbf{147}                                                      & 72                                                     & 145                                                      & 194                                                      & \underline{219}  & 27 & \textbf{53} & \textbf{63} & \underline{69} \\

\multicolumn{1}{c|}{\textit{C-RNN+OMP+LI}} & 44                                                      & \underline{89}                                                        & \textbf{116}                                                       & \textbf{142}                                                       & 115                                                          & 156                                                            & 204                                                           & 251                                                           & \underline{48}                                                     & \underline{95}                                                       & \underline{121}                                                      & \textbf{147}                                                      & 77                                                     & 152                                                      & 195                                                      & \underline{219} & \underline{26} & \textbf{53} & \textbf{63} & \textbf{68} \\ \hline

Object  motion  & \multicolumn{4}{c}{Passing objects}                                                                                                                                                                                                                & \multicolumn{4}{c}{Grasping objects}                                                                                                                                                                                                    & \multicolumn{4}{c}{Cutting food}                                                                                                                                                                                                               & \multicolumn{4}{c}{Mixing objects}                                                                                                                                                                                                             & \multicolumn{4}{c}{Cooking}                                
\\ \hline
\multicolumn{1}{c|}{Time (s)}                                                                  & 0.5 & 1 & 1.5 & 2 & 0.5 &1 & 1.5 & 2 & 0.5 & 1 & 1.5 & 2 & 0.5 & 1 & 1.5 & 2 & 0.5 & 1 & 1.5 & 2 \\ \hline
\multicolumn{1}{c|}{ZV}               & \underline{48}                                                          & 118                                                            & 181                                                            & 237                                                            & 65                                                          & 152                                                            & 226                                                           & 289                                                           & \textbf{29}                                                         & 70                                                           & 104                                                           & 132                                                           & 50                                                         & 126                                                           & 188                                                           & 229  & 16 & \textbf{33} & \textbf{44} & \textbf{53} \\ 
\multicolumn{1}{c|}{RNN}                & 49                                                          & 107                                                            & 154                                                            & 198                                                            & 64                                                          & 139                                                            & 201                                                           & 257                                                           & \textbf{29}                                                         & 70                                                           & 105                                                           & 134                                                           & \underline{47}                                                         & 113                                                           & 166                                                           & 199                                            & 17 & 36 & 48 & 58 \\ \hline

\multicolumn{1}{c|}{\textit{C-RNN+OMP}} & \textbf{44} & \underline{92} & \underline{122} & \underline{150} & \textbf{55} & \textbf{103} & \textbf{136} & \textbf{167} & \underline{31} & \underline{64} & \underline{83} & \underline{97} & \textbf{29} & \textbf{65} & \textbf{90} & \textbf{110} & \textbf{15} & \textbf{33} & \underline{46} & 56 \\ 

\multicolumn{1}{c|}{\textit{C-RNN+OMP+LI}} & \textbf{44}                                                          & \textbf{91}                                                            & \textbf{119}                                                             & \textbf{142}                                                            &                                                          \underline{58} & \underline{112}                                                            & \underline{152}                                                           & \underline{186}                                                           & \textbf{29}                                                         & \textbf{62}                                                           & \textbf{81}                                                           & \textbf{92}                                                           & 51                                                          & \underline{106}                                                           & \underline{145}                                                           & \underline{171}                                                & \textbf{16} & \underline{34} & \underline{46} & \underline{55} \\ \hline




\end{tabular}
}
\vspace{-0.2cm}
\caption{\small{{\bf Class-specific models results.} In this table, every action is independently trained. The results report the mean Euclidean error (in mm), for the 2s prediction of the human motion (top) and object motion (bottom). In all cases, 1s of past observations is provided. The context-based models we propose in this paper are those with the suffixes ``OMP'' and ``LI''. They provide the best results in most sequences.}}
\label{table:result_omp}
\vspace{-0.3cm}
\end{table*}

\begin{table*}
\centering
\footnotesize
\rowcolors{1}{white}{rowblue}
\resizebox{\textwidth}{!}{%
\begin{tabular}{ccccccccccccccccccccc}
\hline
\textbf{All} & \multicolumn{20}{c}{Human motion prediction}                                                                         \\ \hline
Time (s)           & 0.1 & 0.2 & 0.3 & 0.4 & 0.5 & 0.6 & 0.7 & 0.8 & 0.9 & 1   & 1.1 & 1.2 & 1.3 & 1.4 & 1.5 & 1.6 & 1.7 & 1.8 & 1.9 & 2   \\ \hline
ZV~\cite{martinez2017human}        & 24  & 46  & 67  & 87  & 106 & 125 & 143 & 160 & 176 & 190 & 205 & 219 & 231 & 244 & 256 & 267 & 279 & 290 & 300 & 310 \\
RNN~\cite{martinez2017human}        & 29  & 44  & 57  & 68  & 78  & 87  & 96 & 104 & 113 & 121 & 128 & 136 & 143 & 150 & 157 & 164 & 171 & 177 & 184 & 191 \\ \hline

\textit{C-RNN}  & 27    & 46    & 58    & 69    &  79   & 87    &  96   & 104    & 113    & 121    & 129    & 137    & 144    & 152    & 160    & 166    & 174    & 181    & 188    & 196 \\
\textit{C-RNN+OMP} & 46    & 83    & 76    & 82    & 87    & 95    & 101    & 108    & 116    & 123    & 131    & 138    & 146    & 153    & 160    & 167    & 174    & 182    & 189    & 197    \\ \hline
\textit{C-RNN+LI}  & \textbf{21}    & \textbf{39}    & \textbf{52}    & \textbf{63}    & \textbf{72}    & \textbf{80}    & \textbf{89}    & \textbf{97}    & \textbf{104}    & \textbf{111}    & \textbf{118}    & \textbf{125}    & \textbf{131}    & \textbf{137}    & \textbf{144}    & \textbf{150}    & \textbf{157}    & \textbf{163}    & \textbf{170}    & \textbf{177}    \\
\textit{C-RNN+OMP+LI} & 39    & 77    & 77    & 76    & 80    & 87    & 94    & 101    & 108    & 114    & 120    & 126    & 133    & 139    & 145    & 151    & 158    & 165    &    171 & 178    \\ \hline

\textbf{All} & \multicolumn{20}{c}{Object motion prediction}                                                                        \\ \hline
ZV        & \textbf{13}  & \textbf{25}  & \textbf{35}  & \textbf{44}  & 52  & 60  & 68  & 77  & 84  & 90  & 96  & 102 & 109 & 115 & 120 & 125 & 131 & 135 & 140 & 144 \\
RNN         & 15    & 28   & 38   & 46    & 53    & 60    & 68    &     74    & 80    & 85    & 91    & 97    & 102    & 107    & 112    & 117    & 121    & 125    & 130 & 135    \\ \hline
\textit{C-RNN+OMP} & 15    & 26    & 36    & \textbf{44}    & \textbf{50}    & \textbf{55}    & \textbf{61}    & \textbf{67}    & \textbf{73}    & \textbf{79}    & \textbf{84}    & 89    & 94   & 99    & 104    & 108    & 113    & 117    & 121 & 125    \\
\textit{C-RNN+OMP+LI} & 16    & 29    & 39    & 46    & 52    & 57    & 63    & 69    & 75    & \textbf{79}    & \textbf{84}    & \textbf{88}    & \textbf{93}    & \textbf{97}    & \textbf{101}    & \textbf{105}    & \textbf{110}    & \textbf{114}    &    \textbf{117} & \textbf{121}    \\ \hline
\end{tabular}
}
\vspace{-0.2cm}
\caption{\small{{\bf Training with all actions simultaneously.} For each method we train a single model using all actions simultaneously. See also caption in Table~\ref{table:result_omp}.}}
\label{table:result_omp_all}
\vspace{-0.40cm}
\end{table*}

\vspace{-0.2cm}
\section{Implementation details}
\vspace{-0.2cm}

Our model builds on the residual architecture of Martinez~\etal~\cite{martinez2017human} to allow an unbiased comparison with their work. The size of the human and object RNN hidden representations are 1024 and 256, respectively.

After the motion seed, we sample an observation every 100 ms. In all experiments, we encode and decode 10 (1 sec.) and 20 frames (2 sec.) respectively. Larger encoding times did not help in improving the results and significantly increased training time. We augment the train set through random rotation over the height $Z$ in the range $(-180,180]^{\circ}$ and random translation  $X, Y \in (-1500, 1500)$mm. 

We use a similar approach as in~\cite{qi2018learning} to obtain the adjacency matrix. We build a 4D matrix $A$ such that $A_{ij}$ contains the hidden representations $[H_i^t; H_i^t - H_j]$ of nodes $i$ and $j$, extending over the channel dimension. The function  $g(\cdot)$ is formed by two Convolutional Layers of output kernel size $1$ to make computation faster. We do not use bias term in these Convolutional layers nor in the Edge Convolutions. Object representations are formed first by the bounding box position, defined by the minimum and maximum 3D Cartesian points. 

We train the model to minimize $L2$ distance between the predicted and the actual future motion $ \mathcal{L} =  || M(P_{t_o:t-1}) - P_{t:t_f} ||_2 $. The model is trained until convergence, using Adam~\cite{kingma2014adam} with learning rate of 0.0005, beta1 0.5, beta2 0.99 and batch size 16.

\begin{figure*}[t!]
\vspace{-0.4em}
    \includegraphics[trim={0.5cm 0.4cm 1cm 0.85cm}, clip = true,scale=0.63]{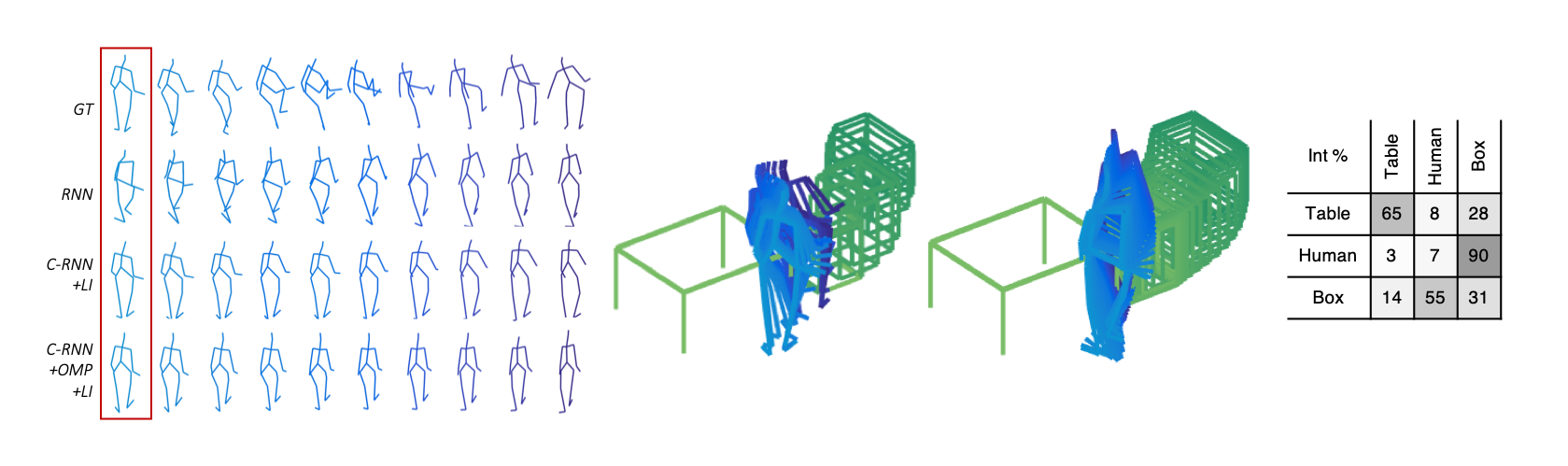}\vspace{-1mm}\\
    \includegraphics[trim={0.5cm 0.6cm 0 1.1cm}, clip = true,scale=0.63]{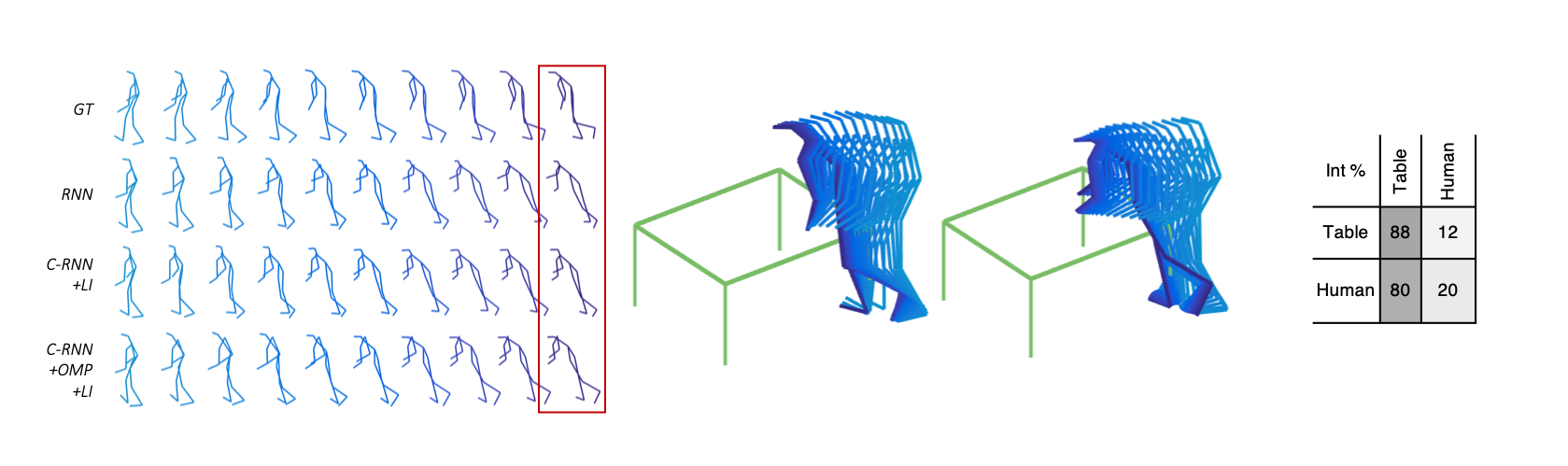}\vspace{-2mm}\\
    \includegraphics[trim={0.5cm 0.6cm 0 0.9cm}, clip = true,scale=0.63]{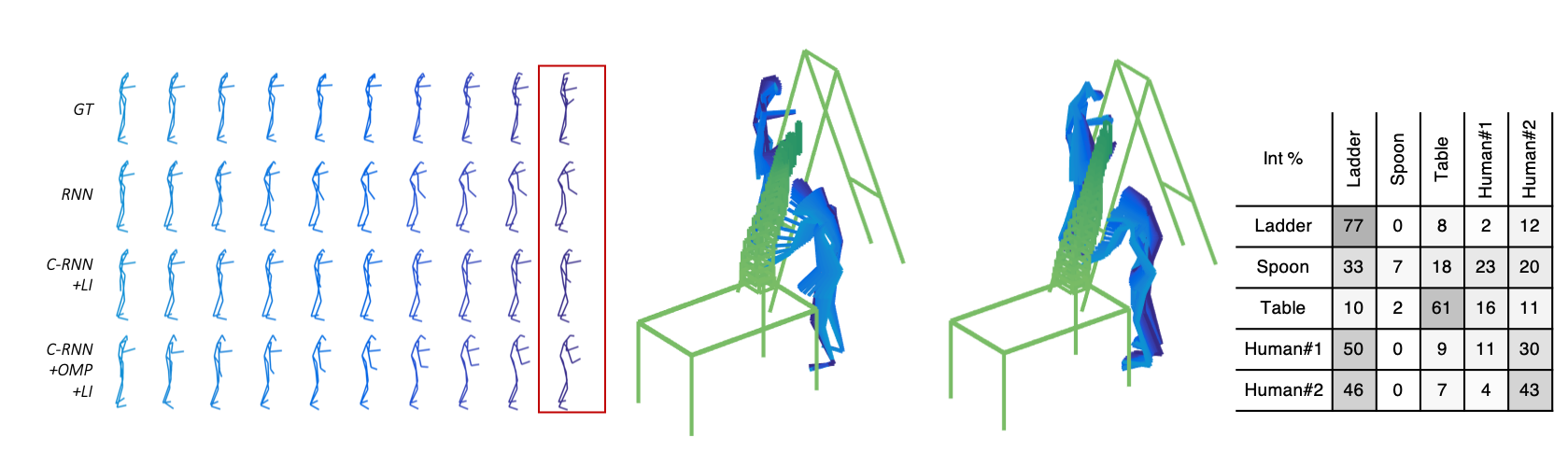}\\
    \put(80, 1){\footnotesize{Joint sequences}}
    \put(200, 1){\footnotesize{Groundtruth motion}}
    \put(290, 1){\footnotesize{\textit{C-RNN+OMP+LI} motion}}
    \put(425, 1){\footnotesize{Predicted $\tilde{A}$}}
    \vspace{-0.1cm}
	\caption{\small{{\bf Qualitative motion generation up to two seconds.} {\bf Left:} Predicted sample frames of our approaches and the baselines. {\bf Center:} Detail of the predictions obtained with our approaches, compared with the ground truth.  Human and object motion are represented from light blue to dark blue and light green to dark green, respectively. Actions, from top to bottom are: A human supports on a table to kick a box, human leaning on a table, and two people (one of them standing on a ladder) passing an object. {\bf Right:} Predicted adjacency matrices representing the interactions learned by our model. Note that these relations are directional (\eg in the last example the ladder highly influences the motion of the Human\#1  (50\%) but the human has little influence over the ladder (11\%).
	Best viewed in color with zoom.}}
	\medskip
	\label{fig:qual_results}
	\vspace{-0.6cm}
\end{figure*}

\vspace{-0.2cm}
\section{Experiments}\label{sec:exp}

\subsection{Preliminaries}

\vspace{1mm} \noindent{\bf Datasets.} 
Large-scale MoCap datasets~\cite{koppula2013learning, h3.6m, cmu_motion_database} provide annotations on the  human poses but do not give any annotation about objects of the scene or any  relevant context information. Therefore, most recent works on human motion prediction are evaluated  without considering context information. Martinez~\etal~\cite{martinez2017human} show that for certain cases, even a simple  zero-velocity baseline may yield better results than  context-less learning models.

To demonstrate  the merits of our approach, we leverage on the Whole-Body Human Motion (WBHM)   Database~\cite{kit_dataset}, a large-scale publicly available dataset containing 3D raw data of multiple individuals and objects. In particular, we use all the activities where human joints are provided and include at least a table. This results in  190 videos and 198K frames,  and a total of 15 tracked object classes. We use the raw recordings Vicon files at 100 Hz to obtain the bounding box of each object in each frame, and select 18 joints  to represent the human skeleton.

We extract different actions representing different levels of complexity on the contextual information. The statistics of this dataset are the following:
 \begin{table}[h!]
\centering
\footnotesize
\rowcolors{1}{white}{rowblue}
\resizebox{\linewidth}{!}{%
\begin{tabular}{ccccccc}
       & \multicolumn{1}{c}{\begin{tabular}[c]{@{}c@{}}Passing\\ objects\end{tabular}} & \multicolumn{1}{c}{\begin{tabular}[c]{@{}c@{}}Grasping/\\ leaving\\ object\end{tabular}} & \multicolumn{1}{c}{\begin{tabular}[c]{@{}c@{}}Cutting\\ food\end{tabular}} & \multicolumn{1}{c}{\begin{tabular}[c]{@{}c@{}}Mixing\\ objects\end{tabular}} & \multicolumn{1}{c}{Cooking} & \multicolumn{1}{c}{All} \\ \hline
\multicolumn{1}{c}{\# objects} & 4                                                                              & 5                                                                                                       & 6                                                                           & 9                                                                             & 12                           & 15                            \\ 
\multicolumn{1}{c}{\# people}                                                     & 2                                                                              & 1                                                                                                       & 1                                                                           & 1                                                                             & 1                            & 1/2                           \\ 
\multicolumn{1}{c}{\# videos}                                                     & 18                                                                             & 36                                                                                                      & 10                                                                          & 17                                                                            & 35                           & 190                           \\ 
\multicolumn{1}{c}{\# frames}                                                     & 30k                                                                            & 31k                                                                                                     & 11k                                                                         & 14k                                                                           & 54k                          & 198k                          \\ \hline
\end{tabular}
}
 
\label{table:data_stats}
\vspace{-0.0cm}
\end{table}

We will report results on   both  action-specific models and also on models trained with  the entire dataset. 


We also run   experiments on the CMU Mocap Database~\cite{cmu_motion_database}. We select the actions that include two people interacting, which include 34 videos with different activities like dancing, talking with hand gestures or boxing. In this case, the objects are not annotated, but we will show that context information from the two users is  useful to improve over context-less models. 

\vspace{1mm} \noindent{\bf Baselines.} We compare our models to the context-less models proposed in~\cite{martinez2017human}. First, we consider the basic residual RNN. We also consider a Zero-Velocity (ZV) baseline that constantly predicts the last observed frame. We also compare to QuaterNet~\cite{pavllo2018quaternet} using their available code, to predict absolute motion prediction. For object motion prediction, we also use a ZV and RNN models~\cite{martinez2017human}, where the position of  an object  is defined by its 3D bounding box.


\vspace{1mm} \noindent{\bf Our models.} We run our context-aware models (\textit{C-RNN}), incrementally adding the main ideas described in the paper.

The basic \textit{C-RNN} in our experiments uses the spatial heuristic described in Section~\ref{sec:context_branch} where interactions depend only on the distance between objects. This model processes context during the past frames, and then uses the last hidden state of the human node for human motion prediction at each step. This is extended by additionally predicting object motion~(\textit{OMP}) and recomputing object interaction from the previous assumption on the predicted positions.
We then evaluate the efficiency of our model for learning interactions (\textit{LI}). Like in  the previously defined experiments, we evaluate a model that considers past contextual information and a model that prolongs object analysis into the future.

\begin{figure*}
\vspace{-0.7em}
\centering
\noindent
\resizebox{0.31\textwidth}{0.30\textwidth}{
\begin{tikzpicture}
\begin{axis}[
    xlabel={Time (sec)},
    ylabel={Interaction \%},
    xmin=0.1, xmax=0.9,
    ymin=0, ymax=100,
    legend pos=north west,
    ymajorgrids=true,
    grid style=dashed,
    legend style={font=\footnotesize},
]
\tikzstyle{every node}=[font=\footnotesize]
\addplot[color=blue, line width=0.5mm]
    coordinates {(0.1,11.666163802146912)(0.2,40.03303349018097)(0.3,38.8535350561142)(0.4,38.63306939601898)(0.5,38.86605501174927)(0.6,38.96074593067169)(0.7,38.3391410112381)(0.8,37.5466912984848)(0.9,36.7510697444916)};
\addplot[color=red, line width=0.5mm]
	coordinates {(0.1,38.33616077899933)(0.2,6.1662159860134125)(0.3,0.18545218044891953)(0.4,0.05163023015484214)(0.5,0.04469105042517185)(0.6,0.061513110995292664)(0.7,0.09484661859460175)(0.8,0.1338200643658638)(0.9,0.16027261735871434)};
\addplot[color=orange, line width=0.5mm]
	coordinates {(0.1,37.4313086271286)(0.2,4.808458685874939)(0.3,0.1622103271074593)(0.4,0.03742173139471561)(0.5,0.026797762257046998)(0.6,0.03324505814816803)(0.7,0.04967434797435999)(0.8,0.07073070155456662)(0.9,0.08771371212787926)};
\addplot[color=purple, line width=0.5mm]
	coordinates {(0.1,50.33883452415466)(0.2,52.504122257232666)(0.3,60.33680438995361)(0.4,60.73930263519287)(0.5,59.962666034698486)(0.6,58.82772207260132)(0.7,58.4588885307312)(0.8,58.68549942970276)(0.9,59.355002641677856)};
\addplot[color=black, line width=0.5mm]
	coordinates {(0.1,67.77819395065308)(0.2,54.52430248260498)(0.3,59.320324659347534)(0.4,59.454894065856934)(0.5,58.29795002937317)(0.6,56.86240792274475)(0.7,55.78961372375488)(0.8,55.119699239730835)(0.9,54.
80384826660156)};
    \legend{Human to Sponge, Sponge to Human, Sponge to Table, Table to Sponge, Table to Human}
\end{axis}
\end{tikzpicture}}
\resizebox{0.31\textwidth}{0.3\textwidth}{
\begin{tikzpicture}
\begin{axis}[
    xlabel={Time (sec)},
    xmin=0.1, xmax=0.9,
    ymin=0, ymax=100,
    legend pos=north west,
    ymajorgrids=true,
    grid style=dashed,
    legend style={font=\footnotesize},
]
\tikzstyle{every node}=[font=\footnotesize]
\addplot[color=blue, line width=0.5mm]
    coordinates {(0.1,41.51618182659149)(0.2,50.446999073028564)(0.3,60.99831461906433)(0.4,74.89581108093262)(0.5,80.24991750717163)(0.6,80.78360557556152)(0.7,78.86382341384888)(0.8,76.13820433616638)(0.9,73.92876148223877)};
\addplot[color=red, line width=0.5mm]
	coordinates {(0.1,7.144319266080856)(0.2,32.76135325431824)(0.3,32.67323672771454)(0.4,29.28035855293274)(0.5,28.882113099098206)(0.6,29.12396490573883)(0.7,29.222694039344788)(0.8,28.98363769054413)(0.9,28.725120425224304)};
\addplot[color=black, line width=0.5mm]
	coordinates {(0.1,52.091825008392334)(0.2,32.90095925331116)(0.3,32.75991082191467)(0.4,27.64458954334259)(0.5,21.839623153209686)(0.6,18.781141936779022)(0.7,17.877237498760223)(0.8,18.320506811141968)(0.9,19.38801258802414)};
\addplot[color=green, line width=0.5mm]
	coordinates {(0.1,3.0231844633817673)(0.2,27.157527208328247)(0.3,23.93999844789505)(0.4,18.054287135601044)(0.5,14.604242146015167)(0.6,12.83574253320694)(0.7,12.369663268327713)(0.8,12.867526710033417)(0.9,13.778959214687347)};
    \legend{Box to Human, Cup to Human, Table to Box, Human to Box}
\end{axis}
\end{tikzpicture}}
\resizebox{0.31\textwidth}{0.3\textwidth}{
\begin{tikzpicture}
\begin{axis}[
    xlabel={Time (sec)},
    xmin=0.1, xmax=0.9,
    ymin=0, ymax=100,
    legend pos=north west,
    ymajorgrids=true,
    grid style=dashed,
    legend style={font=\footnotesize},
]
\tikzstyle{every node}=[font=\footnotesize]
\addplot[color=blue, line width=0.5mm]
    coordinates {(0.1,12.5749871134758)(0.2,10.776278376579285)(0.3,11.366748064756393)(0.4,11.1590676009655)(0.5,10.599587112665176)(0.6,9.98152270913124)(0.7,9.555094689130783)(0.8,9.308766573667526)(0.9,9.214506298303604)(1.0,9.207040071487427)(1.1,9.227833151817322)(1.2,9.272650629281998)(1.3,9.279033541679382)(1.4,9.244327247142792)(1.5,9.245238453149796)(1.6,9.280085563659668)(1.7,9.325356036424637)(1.8,9.37875360250473)(1.9,9.45236012339592)(2.0,9.532910585403442)(2.1,9.606488794088364)(2.2,9.669291228055954)(2.3,9.728856384754181)(2.4,9.792380034923553)(2.5,9.855709969997406)(2.6,9.912393242120743)(2.7,9.963881224393845)(2.8,10.014788806438446)};
\addplot[color=red, line width=0.5mm]
	coordinates {(0.1,13.86386752128601)(0.2,11.98258176445961)(0.3,13.370642066001892)(0.4,13.785970211029053)(0.5,13.815079629421234)(0.6,13.80804181098938)(0.7,13.883021473884583)(0.8,13.962367177009583)(0.9,14.023390412330627)(1.0,14.062973856925964)(1.1,14.025148749351501)(1.2,13.97501677274704)(1.3,13.964737951755524)(1.4,13.965772092342377)(1.5,13.98368775844574)(1.6,14.015702903270721)(1.7,14.024731516838074)(1.8,14.006179571151733)(1.9,13.981945812702179)(2.0,13.968488574028015)(2.1,13.97044062614441)(2.2,13.986586034297943)(2.3,14.010308682918549)(2.4,14.035256206989288)(2.5,14.060381054878235)(2.6,14.086596667766571)(2.7,14.114426076412201)(2.8,14.14170116186142)};
\addplot[color=orange, line width=0.5mm]
	coordinates {(0.1,2.91315671056509)(0.2,20.487438142299652)(0.3,22.244572639465332)(0.4,21.461571753025055)(0.5,21.19763046503067)(0.6,20.841732621192932)(0.7,21.342390775680542)(0.8,22.149905562400818)(0.9,22.801372408866882)(1.0,23.1172576546669)(1.1,23.248785734176636)(1.2,24.734652042388916)(1.3,25.002261996269226)(1.4,24.704186618328094)(1.5,24.497830867767334)(1.6,24.349917471408844)(1.7,24.
23347234725952)(1.8,24.181190133094788)(1.9,24.11370426416397)(2.0,23.92745018005371)(2.1,23.687216639518738)(2.2,23.414212465286255)(2.3,23.091129958629608)(2.4,22.7932870388031)(2.5,22.538480162620544)(
2.6,22.292044758796692)(2.7,22.00917750597)(2.8,21.697548031806946)};
\addplot[color=purple, line width=0.5mm]
	coordinates {(0.1,26.795268058776855)(0.2,40.04674851894379)(0.3,47.564250230789185)(0.4,47.63832688331604)(0.5,46.827369928359985)(0.6,45.66585421562195)(0.7,45.195019245147705)(0.8,45.16726732254028)(0.9,45.21989822387695)(1.0,45.18669247627258)(1.1,45.35430669784546)(1.2,45.853325724601746)(1.3,45.82740664482117)(1.4,45.532673597335815)(1.5,45.30927240848541)(1.6,45.33202350139618)(1.7,45.45798599720001)(1.8,45.593467354774475)(1.9,45.715293288230896)(2.0,45.81235647201538)(2.1,45.94108462333679)(2.2,46.0946649312973)(2.3,46.213555335998535)(2.4,46.31659388542175)(2.5,46.428608894348145)(2.6,46.56125009059906)(2.7,46.70022130012512)(2.8,46.844857931137085)};
\addplot[color=black, line width=0.5mm]
	coordinates {(0.1,25.874072313308716)(0.2,74.57050085067749)(0.3,72.50599265098572)(0.4,68.062824010849)(0.5,72.38926887512207)(0.6,71.73517346382141)(0.7,71.86388373374939)(0.8,71.6002881526947)(0.9,71.62138819694519)(1.0,71.30507826805115)(1.1,70.81896662712097)(1.2,71.30073308944702)(1.3,71.66199684143066)(1.4,71.69363498687744)(1.5,71.57353162765503)(1.6,71.46568894386292)(1.7,71.47923707962036)(1.8,71.710866689682)(1.9,71.61923050880432)(2.0,71.48324251174927)(2.1,71.50300741195679)(2.2,71.61378264427185)(2.3,71.5330421924591)(2.4,71.52886986732483)(2.5,71.55715823173523)(2.6,71.64846062660217)(2.7,71.6801404953003)(2.8,71.72592878341675)};
    \legend{Knife to Knife, Bottle to Bottle, Human to Human, Table to Table, Ladder to Ladder}
\end{axis}
\end{tikzpicture}}\\
\vspace{-0.4cm}
\caption{\small{{\bf Average interactions refined by the model during the past observations of the context}. In the left and center plots, we depict relevant interactions for {\em table cleaning} and {\em moving box} activities respectively. In the first case, notice the table affects significantly the sponge and human, which initially moves towards the table to clean it. Similarly, in the second case, the human moves towards a box on the ground,   picks it up and puts it  on the table.
The right plot shows average self-interaction percentages among all the test samples, for relevant object types. We found that non-moving objects like tables or ladders consistently have very little influence from other objects. Likewise, passive objects that are often moved by a human, such as knives or bottles, are more influenced by them and leave self-influence relatively low.} }
\label{fig:interactions_statistics}
\vspace{-0.4cm}

\end{figure*}
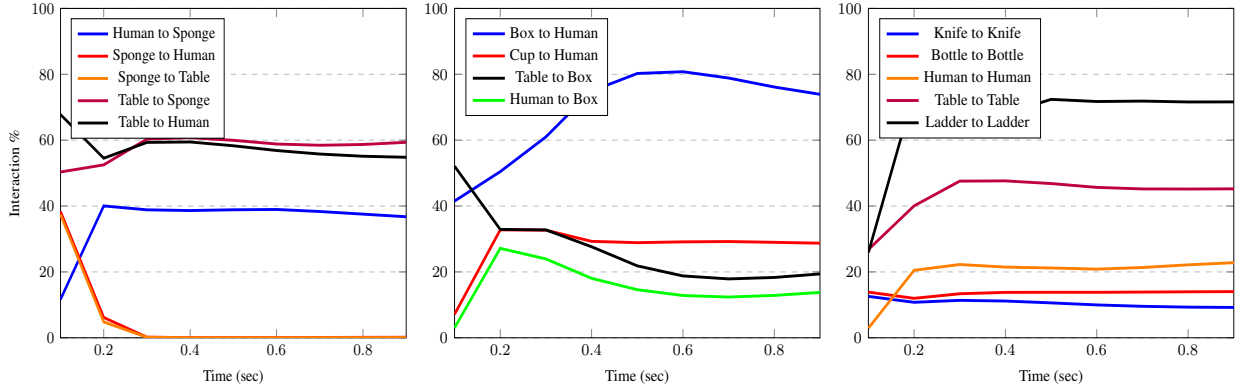

\vspace{1mm} \noindent{\bf Evaluation metric.} 
Previous works on human motion prediction focus mainly on predicting relative motion~\cite{martinez2017human, gui2018adversarial, pavllo2018quaternet}, using joint angles. However, our model reasons about the full scene and is able to predict absolute motion in Cartesian coordinates. Therefore, we use the  mean Euclidean Distance  (in mm)  between predictions and real future motion, obtained from the unnormalized predictions in the 3D space. For human motion prediction, we take into account the 18 joints defined in the human skeleton. For objects, we consider the eight 3D vertices of their bounding boxes.

\subsection{Results on the WBHM  Dataset}

\vspace{1mm} \noindent{\bf Quantitative results.} Table~\ref{table:result_omp} summarizes the performance of class-specific models trained on different activities. 
Table~\ref{table:result_omp_all} provides results at much higher temporal resolution for     models trained using all the dataset, reporting the mean Euclidean distance between predictions and ground truth every 100 ms. In all cases, 1 second of past observations is provided and 2 seconds are predicted.

The performance of models that consider a threshold-based binary interaction vary significantly between classes, suggesting they are effectively unable to understand the context as  done by models that learn the actual interactions~(\textit{LI}). Notice that even the  basic \textit{C-RNN} does not yet provide a consistent improvement compared to state-of-art models. The same model that additionally learns interactions (C-RNN+LI) obtains a significant boost in most cases. Nonetheless, activities such as passing objects or grasping require attending to items that are at variable distances.

Regarding the complexity of the scene, most improvement comes from scenes with a small number of objects where interactions are well defined and actions are more predictable. For cooking activities, there are several objects in a table next to the human. 
Different motion options are possible and, as uncertainty grows, 
the model seems unable to confidently understand interactions. Because of this, context-aware models do not provide such a significant improvement as in previous activities.
Considering all actions simultaneously seems to favor even more the context-aware approaches and, specially, those that learn interactions ({\em C-RNN+LI}  and {\em C-RNN+OPM+LI}).

\begin{figure}[t!]
\includegraphics[trim={0.8cm 0.8cm 0cm 1cm}, clip = true, width=0.235\textwidth]{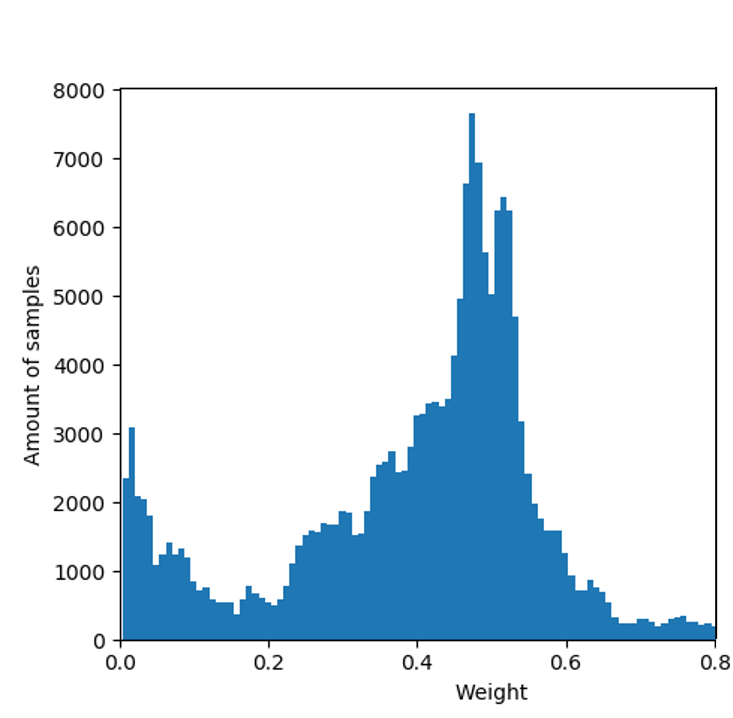}
\includegraphics[trim={0.8cm 0.8cm 0cm 1cm}, clip = true, width=0.235\textwidth]{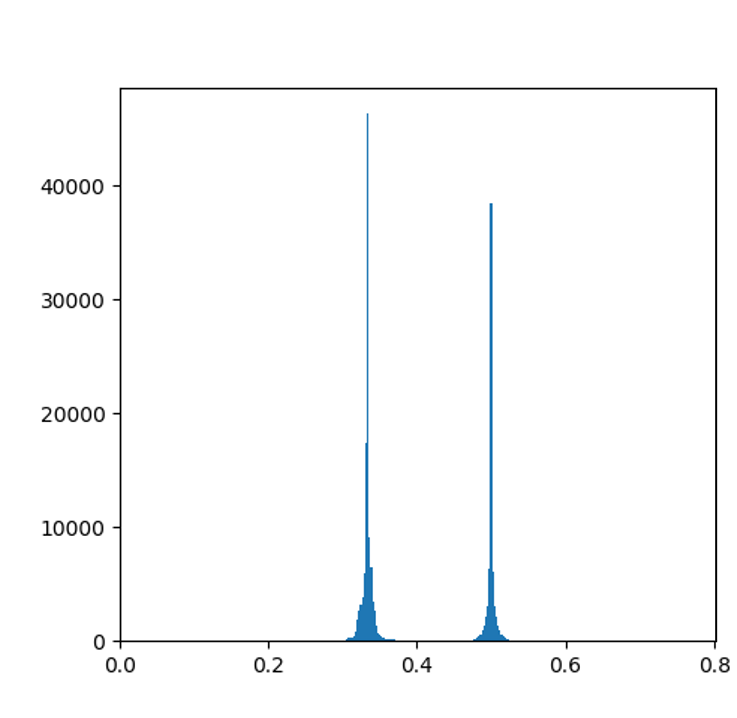}\\
\put(40, 5){\tiny{Interaction weight (EC)}}
\put(150, 5){\tiny{Interaction weight (GAT)}}
\\
\vspace{-0.8cm}
\caption{\small{{\bf Interaction strength histogram predicted by EC and GAT models.} These include interactions predicted among all humans and objects after two observations are given to the networks. For simplicity, we depict the histogram from tasks whose context only contain two or three nodes. On the left, interactions learnt by EC-based model, spanning a wide range of values, up to interaction strengths of more than 80\%. On the right, GAT-based model, which predicts all interaction weights similarly and therefore we can only see peaks at $1/2$ and $1/3$. }}
\label{fig:interactions_architectures}
\vspace{-0.2cm}
\end{figure}

\begin{table*}
\vspace{-2mm}
\centering
\footnotesize
\rowcolors{1}{white}{rowblue}
\resizebox{1.0\textwidth}{!}{%
\begin{tabular}{c|cccc|cccc|cccc|cccc}
\hline
& \multicolumn{4}{c|}{Noise-free input} & \multicolumn{12}{c}{ Noisy input (HMP)} \\
Model &  \multicolumn{4}{c|}{HMP Models}  & \multicolumn{4}{c|}{25 mm} & \multicolumn{4}{c|}{50 mm} & \multicolumn{4}{c}{ 100 mm} \\
Time (s) & 0.5 & 1 & 1.5 & 2 & 0.5 & 1 & 1.5 & 2 & 0.5 & 1 & 1.5 & 2 & 0.5 & 1 & 1.5 & 2\\
\hline
ZV \cite{martinez2017human} & 61 & 150 & 223 & 281 & 70 & 156 & 227 & 284 & 87 & 168 & 236 & 292 & 122 & 195 & 260 & 313 \\
RNN \cite{martinez2017human} & 55 & 110 & 148 & 179 & 72 & 122 & 156 & 185 & 83 & 127 & 162 & 192 & 112 & 184 & 214 & 226 \\ \hline
C-RNN+LI & \underline{52} & 104 & \textbf{136} &\textbf{161} & \textbf{58} & \textbf{107} & \textbf{139} & \textbf{166} & \textbf{69} & \textbf{113} & \underline{147} & \textbf{175} & \textbf{99} & 136 & 175 & 208 \\
C-RNN+OMP+LI & 56 & 109 & 140 & \underline{165} & 62 & 111 & 142 & \underline{167} & \underline{71} & \underline{116} & \textbf{146} & \underline{171} & \underline{103} & \underline{132} & \textbf{162} & \textbf{187} \\ \hline

Model &  \multicolumn{4}{c|}{OMP Models} & \multicolumn{4}{c}{25 mm} & \multicolumn{4}{c}{50 mm} & \multicolumn{4}{c}{ 100 mm} \\
\hline
ZV & 42 & 100 & 149 & 188 & 53 & 106 & 154 & 191 & \textbf{66} & 118 & 164 & 199 & 106 & 151 & 187 & 223 \\
RNN & 41 & 93 & 135 & 169 & 52 & 99 & 141 & 174 & 69 & 113 & 151 & 183 & \textbf{105} & 142 & 181 & 208 \\ 
C-RNN+OMP+LI & \textbf{40} & \textbf{81} & \textbf{109} & \textbf{129} & \textbf{51} & \textbf{88} & \textbf{115} & \textbf{134} & 67 & \textbf{100} & \textbf{126} & \textbf{144} & 106 & \textbf{132} & \textbf{156} & \textbf{172} \\ \hline
\end{tabular}}
\vspace{-0.1cm}
\caption{\small{{\bf Robustness to noise in Human and Object Motion Prediction.} Average performance of the principal models when using the original test set (Noise-free input), compared to their performance when seeing noisy observations.}}
\label{table:noisy_observations}
\vspace{-0.2cm}
\end{table*}

\vspace{1mm} \noindent{\bf Qualitative results.} Figure~\ref{fig:qual_results}-left shows the motion generation results   of our two main  models, compared to the   baseline~\cite{martinez2017human} on different classes. 
We did not include the   Zero-Velocity baseline as it does not provide interesting motion even though it has remained a difficult baseline on uncertain activities.
We have marked some specific frames in which context-aware approaches improve the RNN baseline. 

For human motion prediction, poses generated are frequently more semantically-related to their closest objects than context-less models. 
For instance, as shown in the last action of Figure~\ref{fig:qual_results}, people holding objects tend to move the relevant hand. For object motion prediction, context-less model predictions hardly move from their original position. 

Regarding the interactions predicted by the model, we notice coherent patterns in many activities. For example, drinking videos generate strong Cup-Human relationships. In Figure~\ref{fig:interactions_statistics}, we represent the average predicted interactions for different actions. These are gathered from the \textit{C-RNN+OMP+LI}. This model provides more intense Object-Object interactions than \textit{C-RNN+LI}, which does not need to obtain such meaningful representations for objects as only human contextual representations are  used. Note that the models learn to predict interactions that provide information relevant to future pose, and thus improve motion predictions. Interactions here do not necessarily respond to actual action relationships.

We finally study the effect of the Graph architecture in the learned interactions.
Graph Attention Networks (GATs) and Edge Convolutions (EC) provide an attention mechanism to measure the interaction strength. Nevertheless,   we found that GAT-based networks consider all interactions of similar importance, while EC-based architectures are able to predict a continuous and wide range of attention values. We show this in   Figure~\ref{fig:interactions_architectures}.

\begin{table}
\centering
\footnotesize
\rowcolors{1}{white}{rowblue}
\resizebox{0.5\textwidth}{!}{
\begin{tabular}{ccccccccc}
\hline
\multicolumn{9}{c}{CMU MoCap Dataset} \\
\multicolumn{1}{c|}{Time (s)} & \multicolumn{2}{c}{0.5}  & \multicolumn{2}{c}{1.0} & \multicolumn{2}{c}{1.5} & \multicolumn{2}{c}{2.0} \\  \hline
 & $\mu$ & $\sigma$ & $\mu$ & $\sigma$ & $\mu$ & $\sigma$ & $\mu$ & $\sigma$ \\  \hline

\multicolumn{1}{c|}{\textit{ZV}} & 127 & 32  & 271 & 66 & \underline{374} & 86 & \underline{460} & 97 \\
\multicolumn{1}{c|}{\textit{RNN}} & \underline{125} & 28 & \underline{267} & 58 & 378 & 77 & 477 & 92 \\
\multicolumn{1}{c|}{\textit{QuaterNet}} & 138 & 26 & 279 & 58 & 378 & 82 & 466 & 95 \\
\hline
\multicolumn{1}{c|}{\textit{C-RNN+LI}} & \textbf{124} & 27 & \textbf{257} & 53 & \textbf{352} & 65 & \textbf{435} & 78 \\
\hline
\end{tabular}}
\vspace{-1mm}
\caption{\small{\textbf{Mean and Std of prediction errors (mm) on the CMU dataset.} Our Context-aware model C-RNN+LI outperforms baselines even though context only consists of two people. }}
\label{table:cmu_mocap}
\vspace{-3mm}
\end{table}

\subsection{Results on the CMU MoCap Dataset}


We train the models again on the CMU MoCap Database, obtaining the results depicted in Table~\ref{table:cmu_mocap}. In this setup, the users perform very energetic activities like dancing or boxing, which implies that absolute motion is larger, and error on the CMU MoCap database being in average more than twice that in the former database.
In this case, only two nodes are observed in each video for the two people being tracked. Since no information about actions or objects is given, we do not provide results on \textit{OMP}. However, we find our proposed model \textit{C-RNN+LI} outperforms all other baselines significantly, specially in the long-term.

\subsection{Robustness to noise}
 
 All previous works on human motion prediction use ground truth MoCap data as past observations. Nevertheless, real applications will receive joint observations from \eg human pose estimation models, such as OpenPose~\cite{cao2018openpose} or AlphaPose~\cite{fang2017rmpe, xiu2018pose}, which are prone to suffer from noise and mis-detections, specially under strong occlusions.
In these subsection, we therefore evaluate the resilience of our proposed models and previous baselines to noise in the input observations.
Predictions are evaluated on the original ground truth data.
The 3D coordinates of past observations (both in human and objects positions) are corrupted by additive Gaussian noise $\mathcal{N}(0,\,\sigma^{2})$. In Table ~\ref{table:noisy_observations} we show the results of this experiment, with different values of $\sigma$. Interestingly, the error in the predictions gracefully increases with the noise, but still, our approach performs consistently better than those approaches that do not consider the context information. Indeed, the best context-aware models (\textit{C-RNN+LI} and \textit{C-RNN+OMP+LI}) with noise up to $\sigma=50mm$, perform better than context-less baselines with no noise in the input.

\section{Conclusion}
In this work, we explore a context-aware motion prediction architecture, using a semantic-graph representation where objects and humans are represented by nodes independently of the number of objects or complexity of the environment. We extensively analyze their contribution for human motion prediction. The results observed in different actions suggest that the models proposed are able to understand human activities  significantly better than state-of-art models which do not use context, improving both human and object motion prediction.


\vspace{1mm} \noindent{\bf Acknowledgements}: 
This work has been partially funded by Spanish government with the projects HuMoUR TIN2017-90086-R and  the ERA-Net Chistera project IPALM PCI2019-103386. We also thank Nvidia for hardware donation.

{\small
\bibliographystyle{ieee_fullname}
\bibliography{top.bib}

\begin{thebibliography}{10}\itemsep=-1pt

\bibitem{Aksan_2019_ICCV}
Emre Aksan, Manuel Kaufmann, and Otmar Hilliges.
\newblock Structured prediction helps 3d human motion modelling.
\newblock In {\em ICCV}, 2019.

\bibitem{arjovsky2017wasserstein}
Martin Arjovsky, Soumith Chintala, and L{\'e}on Bottou.
\newblock Wasserstein gan.
\newblock {\em arXiv preprint arXiv:1701.07875}, 2017.

\bibitem{barsoum2018hp}
Emad Barsoum, John Kender, and Zicheng Liu.
\newblock Hp-gan: Probabilistic 3d human motion prediction via gan.
\newblock In {\em CVPR-Workshop}, 2018.

\bibitem{baser2019fantrack}
Erkan Baser, Venkateshwaran Balasubramanian, Prarthana Bhattacharyya, and
  Krzysztof Czarnecki.
\newblock Fantrack: 3d multi-object tracking with feature association network.
\newblock In {\em IV}, 2019.

\bibitem{bogo2016keep}
Federica Bogo, Angjoo Kanazawa, Christoph Lassner, Peter Gehler, Javier Romero,
  and Michael~J Black.
\newblock Keep it smpl: Automatic estimation of 3d human pose and shape from a
  single image.
\newblock In {\em ECCV}, 2016.

\bibitem{byravan2017se3}
Arunkumar Byravan and Dieter Fox.
\newblock Se3-nets: Learning rigid body motion using deep neural networks.
\newblock In {\em ICRA}, 2017.

\bibitem{cao2018openpose}
Zhe Cao, Gines Hidalgo, Tomas Simon, Shih-En Wei, and Yaser Sheikh.
\newblock Openpose: realtime multi-person 2d pose estimation using part
  affinity fields.
\newblock {\em arXiv preprint arXiv:1812.08008}, 2018.

\bibitem{chang2019argoverse}
Ming-Fang Chang, John Lambert, Patsorn Sangkloy, Jagjeet Singh, Slawomir Bak,
  Andrew Hartnett, De Wang, Peter Carr, Simon Lucey, Deva Ramanan, et~al.
\newblock Argoverse: 3d tracking and forecasting with rich maps.
\newblock In {\em Proceedings of the IEEE Conference on Computer Vision and
  Pattern Recognition}, 2019.

\bibitem{chen20153d}
Xiaozhi Chen, Kaustav Kundu, Yukun Zhu, Andrew~G Berneshawi, Huimin Ma, Sanja
  Fidler, and Raquel Urtasun.
\newblock 3d object proposals for accurate object class detection.
\newblock In {\em Advances in Neural Information Processing Systems}, pages
  424--432, 2015.

\bibitem{chen2018graph}
Yunpeng Chen, Marcus Rohrbach, Zhicheng Yan, Shuicheng Yan, Jiashi Feng, and
  Yannis Kalantidis.
\newblock Graph-based global reasoning networks.
\newblock {\em arXiv preprint arXiv:1811.12814}, 2018.

\bibitem{corona2018pose}
Enric Corona, Kaustav Kundu, and Sanja Fidler.
\newblock Pose estimation for objects with rotational symmetry.
\newblock In {\em IROS}, 2018.

\bibitem{corona2020ganhand}
Enric Corona, Albert Pumarola, Guillem Aleny{\`a}, Francesc Moreno-Noguer, and
  Gr{\'e}gory Rogez.
\newblock Ganhand: Predicting human grasp affordances in multi-object scenes.
\newblock In {\em CVPR}, 2020.

\bibitem{fang2017rmpe}
Hao-Shu Fang, Shuqin Xie, Yu-Wing Tai, and Cewu Lu.
\newblock Rmpe: Regional multi-person pose estimation.
\newblock In {\em CVPR}, 2017.

\bibitem{fragkiadaki2015recurrent}
Katerina Fragkiadaki, Sergey Levine, Panna Felsen, and Jitendra Malik.
\newblock Recurrent network models for human dynamics.
\newblock In {\em ICCV}, 2015.

\bibitem{ghosh2017learning}
Partha Ghosh, Jie Song, Emre Aksan, and Otmar Hilliges.
\newblock Learning human motion models for long-term predictions.
\newblock In {\em 2017 International Conference on 3D Vision (3DV)}, 2017.

\bibitem{gkioxari2015actions}
Georgia Gkioxari, Ross Girshick, and Jitendra Malik.
\newblock Actions and attributes from wholes and parts.
\newblock In {\em ICCV}, 2015.

\bibitem{goodfellow2014generative}
Ian Goodfellow, Jean Pouget-Abadie, Mehdi Mirza, Bing Xu, David Warde-Farley,
  Sherjil Ozair, Aaron Courville, and Yoshua Bengio.
\newblock Generative adversarial nets.
\newblock In {\em NIPS}, 2014.

\bibitem{grabner20183d}
Alexander Grabner, Peter~M Roth, and Vincent Lepetit.
\newblock 3d pose estimation and 3d model retrieval for objects in the wild.
\newblock In {\em Proceedings of the IEEE Conference on Computer Vision and
  Pattern Recognition}, pages 3022--3031, 2018.

\bibitem{groueix2018atlasnet}
Thibault Groueix, Matthew Fisher, Vladimir~G. Kim, Bryan Russell, and Mathieu
  Aubry.
\newblock {AtlasNet: A Papier-M\^ach\'e Approach to Learning 3D Surface
  Generation}.
\newblock In {\em CVPR}, 2018.

\bibitem{gui2018adversarial}
Liang-Yan Gui, Yu-Xiong Wang, Xiaodan Liang, and Jos{\'e}~MF Moura.
\newblock Adversarial geometry-aware human motion prediction.
\newblock In {\em ECCV}, 2018.

\bibitem{he2016deep}
Kaiming He, Xiangyu Zhang, Shaoqing Ren, and Jian Sun.
\newblock Deep residual learning for image recognition.
\newblock In {\em CVPR}, 2016.

\bibitem{herzig2018mapping}
Roei Herzig, Moshiko Raboh, Gal Chechik, Jonathan Berant, and Amir Globerson.
\newblock Mapping images to scene graphs with permutation-invariant structured
  prediction.
\newblock In {\em NIPS}, 2018.

\bibitem{Hu_2018_CVPR}
Han Hu, Jiayuan Gu, Zheng Zhang, Jifeng Dai, and Yichen Wei.
\newblock Relation networks for object detection.
\newblock In {\em CVPR}, 2018.

\bibitem{h3.6m}
Catalin Ionescu, Dragos Papava, Vlad Olaru, and Cristian Sminchisescu.
\newblock Human3. 6m: Large scale datasets and predictive methods for 3d human
  sensing in natural environments.
\newblock {\em TPAMI}, 2014.

\bibitem{jain2016structural}
Ashesh Jain, Amir~R Zamir, Silvio Savarese, and Ashutosh Saxena.
\newblock Structural-rnn: Deep learning on spatio-temporal graphs.
\newblock In {\em CVPR}, 2016.

\bibitem{kim2019instance}
Kyung-Rae Kim, Whan Choi, Yeong~Jun Koh, Seong-Gyun Jeong, and Chang-Su Kim.
\newblock Instance-level future motion estimation in a single image based on
  ordinal regression.
\newblock In {\em ICCV}, 2019.

\bibitem{kingma2014adam}
Diederik~P Kingma and Jimmy Ba.
\newblock Adam: A method for stochastic optimization.
\newblock {\em arXiv preprint arXiv:1412.6980}, 2014.

\bibitem{GCNs}
Thomas~N Kipf and Max Welling.
\newblock Semi-supervised classification with graph convolutional networks.
\newblock {\em arXiv preprint arXiv:1609.02907}, 2016.

\bibitem{koppula2013learning}
Hema~Swetha Koppula, Rudhir Gupta, and Ashutosh Saxena.
\newblock Learning human activities and object affordances from rgb-d videos.
\newblock {\em IJRR}, 2013.

\bibitem{koppula2016anticipating}
Hema~S Koppula and Ashutosh Saxena.
\newblock Anticipating human activities using object affordances for reactive
  robotic response.
\newblock {\em TPAMI}, 2016.

\bibitem{kovar2008motion}
Lucas Kovar, Michael Gleicher, and Fr{\'e}d{\'e}ric Pighin.
\newblock Motion graphs.
\newblock In {\em SIGGRAPH}, 2008.

\bibitem{kratzer2019motion}
Philipp Kratzer, Marc Toussaint, and Jim Mainprice.
\newblock Motion prediction with recurrent neural network dynamical models and
  trajectory optimization.
\newblock {\em arXiv preprint arXiv:1906.12279}, 2019.

\bibitem{kundu2018bihmp}
Jogendra~Nath Kundu, Maharshi Gor, and R~Venkatesh Babu.
\newblock Bihmp-gan: Bidirectional 3d human motion prediction gan.
\newblock {\em arXiv preprint arXiv:1812.02591}, 2018.

\bibitem{cmu_motion_database}
CMU~Graphics Lab.
\newblock Cmu motion capture database.
\newblock \url{http://mocap.cs.cmu.edu/}.

\bibitem{lea2016segmental}
Colin Lea, Austin Reiter, Ren{\'e} Vidal, and Gregory~D Hager.
\newblock Segmental spatiotemporal cnns for fine-grained action segmentation.
\newblock In {\em ECCV}, 2016.

\bibitem{li2017situation}
Ruiyu Li, Makarand Tapaswi, Renjie Liao, Jiaya Jia, Raquel Urtasun, and Sanja
  Fidler.
\newblock Situation recognition with graph neural networks.
\newblock In {\em ICCV}, 2017.

\bibitem{li2018transferable}
Yong-Lu Li, Siyuan Zhou, Xijie Huang, Liang Xu, Ze Ma, Hao-Shu Fang, Yan-Feng
  Wang, and Cewu Lu.
\newblock Transferable interactiveness prior for human-object interaction
  detection.
\newblock {\em arXiv preprint arXiv:1811.08264}, 2018.

\bibitem{liu2018structure}
Yong Liu, Ruiping Wang, Shiguang Shan, and Xilin Chen.
\newblock Structure inference net: object detection using scene-level context
  and instance-level relationships.
\newblock In {\em CVPR}, 2018.

\bibitem{ma2018attend}
Chih-Yao Ma, Asim Kadav, Iain Melvin, Zsolt Kira, Ghassan AlRegib, and Hans
  Peter~Graf.
\newblock Attend and interact: Higher-order object interactions for video
  understanding.
\newblock In {\em CVPR}, 2018.

\bibitem{kit_dataset}
Christian Mandery, {\"O}mer Terlemez, Martin Do, Nikolaus Vahrenkamp, and Tamim
  Asfour.
\newblock The kit whole-body human motion database.
\newblock In {\em ICAR}, 2015.

\bibitem{mandikal2019dense}
Priyanka Mandikal and Venkatesh~Babu Radhakrishnan.
\newblock Dense 3d point cloud reconstruction using a deep pyramid network.
\newblock In {\em WACV}, 2019.

\bibitem{Mao_2019_ICCV}
Wei Mao, Miaomiao Liu, Mathieu Salzmann, and Hongdong Li.
\newblock Learning trajectory dependencies for human motion prediction.
\newblock In {\em ICCV}, 2019.

\bibitem{martinez2017human}
Julieta Martinez, Michael~J Black, and Javier Romero.
\newblock On human motion prediction using recurrent neural networks.
\newblock In {\em CVPR}, 2017.

\bibitem{Moreno-Noguer_2017_CVPR}
Francesc Moreno-Noguer.
\newblock 3d human pose estimation from a single image via distance matrix
  regression.
\newblock In {\em CVPR}, 2017.

\bibitem{newell2017pixels}
Alejandro Newell and Jia Deng.
\newblock Pixels to graphs by associative embedding.
\newblock In {\em NIPS}, 2017.

\bibitem{ni2016progressively}
Bingbing Ni, Xiaokang Yang, and Shenghua Gao.
\newblock Progressively parsing interactional objects for fine grained action
  detection.
\newblock In {\em CVPR}, 2016.

\bibitem{paden2016survey}
Brian Paden, Michal {\v{C}}{\'a}p, Sze~Zheng Yong, Dmitry Yershov, and Emilio
  Frazzoli.
\newblock A survey of motion planning and control techniques for self-driving
  urban vehicles.
\newblock {\em IV}, 2016.

\bibitem{parisot2018disease}
Sarah Parisot, Sofia~Ira Ktena, Enzo Ferrante, Matthew Lee, Ricardo Guerrero,
  Ben Glocker, and Daniel Rueckert.
\newblock Disease prediction using graph convolutional networks: Application to
  autism spectrum disorder and alzheimer’s disease.
\newblock {\em Medical image analysis}, 48, 2018.

\bibitem{pavlakos2017coarse}
Georgios Pavlakos, Xiaowei Zhou, Konstantinos~G Derpanis, and Kostas
  Daniilidis.
\newblock Coarse-to-fine volumetric prediction for single-image 3d human pose.
\newblock In {\em CVPR}, 2017.

\bibitem{pavllo2018quaternet}
Dario Pavllo, David Grangier, and Michael Auli.
\newblock Quaternet: A quaternion-based recurrent model for human motion.
\newblock {\em arXiv preprint arXiv:1805.06485}, 2018.

\bibitem{pumarola2020c}
Albert Pumarola, Stefan Popov, Francesc Moreno-Noguer, and Vittorio Ferrari.
\newblock {C-Flow: Conditional Generative Flow Models for Images and 3D Point
  Clouds}.
\newblock In {\em CVPR}, 2020.

\bibitem{pumarola20193dpeople}
Albert Pumarola, Jordi Sanchez-Riera, Gary Choi, Alberto Sanfeliu, and Francesc
  Moreno-Noguer.
\newblock 3dpeople: Modeling the geometry of dressed humans.
\newblock In {\em ICCV}, 2019.

\bibitem{qi2018learning}
Siyuan Qi, Wenguan Wang, Baoxiong Jia, Jianbing Shen, and Song-Chun Zhu.
\newblock Learning human-object interactions by graph parsing neural networks.
\newblock In {\em ECCV}, 2018.

\bibitem{ruiz2019human}
Alejandro~Hernandez Ruiz, Juergen Gall, and Francesc Moreno-Noguer.
\newblock Human motion prediction via spatio-temporal inpainting.
\newblock In {\em ICCV}, 2019.

\bibitem{simo2013joint}
Edgar Simo-Serra, Ariadna Quattoni, Carme Torras, and Francesc Moreno-Noguer.
\newblock A joint model for 2d and 3d pose estimation from a single image.
\newblock In {\em CVPR}, 2013.

\bibitem{simo2012single}
Edgar Simo-Serra, Arnau Ramisa, Guillem Aleny{\`a}, Carme Torras, and Francesc
  Moreno-Noguer.
\newblock Single image 3d human pose estimation from noisy observations.
\newblock In {\em CVPR}, 2012.

\bibitem{simo2017humanpose}
Edgar Simo-Serra, Carme Torras, and Francesc Moreno-Noguer.
\newblock 3d human pose tracking priors using geodesic mixture models.
\newblock 2017.

\bibitem{velivckovic2017graph}
Petar Veli{\v{c}}kovi{\'c}, Guillem Cucurull, Arantxa Casanova, Adriana Romero,
  Pietro Lio, and Yoshua Bengio.
\newblock Graph attention networks.
\newblock {\em arXiv preprint arXiv:1710.10903}, 2017.

\bibitem{vijayanarasimhan2017sfm}
Sudheendra Vijayanarasimhan, Susanna Ricco, Cordelia Schmid, Rahul Sukthankar,
  and Katerina Fragkiadaki.
\newblock Sfm-net: Learning of structure and motion from video.
\newblock {\em arXiv preprint arXiv:1704.07804}, 2017.

\bibitem{wang2014robust}
Chunyu Wang, Yizhou Wang, Zhouchen Lin, Alan~L Yuille, and Wen Gao.
\newblock Robust estimation of 3d human poses from a single image.
\newblock In {\em CVPR}, 2014.

\bibitem{wang2018videos}
Xiaolong Wang and Abhinav Gupta.
\newblock Videos as space-time region graphs.
\newblock In {\em ECCV}, 2018.

\bibitem{EdgeConvs}
Yue Wang, Yongbin Sun, Ziwei Liu, Sanjay~E Sarma, Michael~M Bronstein, and
  Justin~M Solomon.
\newblock Dynamic graph cnn for learning on point clouds.
\newblock {\em arXiv preprint arXiv:1801.07829}, 2018.

\bibitem{xie2019rigid}
Jiayin Xie and Nilanjan Chakraborty.
\newblock Rigid body motion prediction with planar non-convex contact patch.
\newblock In {\em ICRA}, 2019.

\bibitem{xiu2018pose}
Yuliang Xiu, Jiefeng Li, Haoyu Wang, Yinghong Fang, and Cewu Lu.
\newblock Pose flow: Efficient online pose tracking.
\newblock {\em arXiv preprint arXiv:1802.00977}, 2018.

\bibitem{yatskar2016situation}
Mark Yatskar, Luke Zettlemoyer, and Ali Farhadi.
\newblock Situation recognition: Visual semantic role labeling for image
  understanding.
\newblock In {\em CVPR}, 2016.

\bibitem{zhang2019predicting}
Jason~Y Zhang, Panna Felsen, Angjoo Kanazawa, and Jitendra Malik.
\newblock Predicting 3d human dynamics from video.
\newblock In {\em ICCV}, 2019.

\end{thebibliography}
}

\end{document}